\documentclass{article}

\usepackage{arxiv}

\usepackage[utf8]{inputenc} 
\usepackage[T1]{fontenc}    
\usepackage{hyperref}       
\usepackage{url}            
\usepackage{booktabs}       
\usepackage{amsfonts}       
\usepackage{nicefrac}       
\usepackage{microtype}      
\usepackage{lipsum}		
\usepackage{amsmath,amssymb,amsfonts}
\usepackage{graphicx}
\usepackage{natbib}
\usepackage{doi}
\usepackage{subcaption}
\usepackage{multirow}
\usepackage{algorithm}
\usepackage{algpseudocode}
\usepackage{comment}
\usepackage{bm}
\usepackage[table]{xcolor}
\usepackage{float}
\usepackage{placeins}

\title{L2GTX: From Local to Global Time Series Explanations}

\author{
Ephrem Tibebe Mekonnen\thanks{Accepted for publication in the Springer proceedings of the 4th World Conference on Explainable Artificial Intelligence (xAI-2026).}\\
School of Computer Science, Technological University Dublin\\
Dublin, Ireland\\
Artificial Intelligence and Cognitive Load Research Lab, University College Cork\\
Cork, Ireland\\
\texttt{D22125038@mytudublin.ie}
\And
Luca Longo\\
School of Computer Science and Information Technology, University College Cork\\
Cork, Ireland\\
Artificial Intelligence and Cognitive Load Research Lab, University College Cork\\
Cork, Ireland\\
\texttt{luca.longo@ucc.ie}
\And
Lucas Rizzo\\
School of Computer Science, Technological University Dublin\\
Dublin, Ireland\\
Artificial Intelligence and Cognitive Load Research Lab, University College Cork\\
Cork, Ireland\\
\texttt{lucas.rizzo@tudublin.ie}
\And
Pierpaolo Dondio\\
School of Computing, Dublin City University\\
Dublin, Ireland\\
\texttt{pierpaolo.dondio@dcu.ie}
}
\date{}

\renewcommand{\headeright}{}
\renewcommand{\undertitle}{}
\renewcommand{\shorttitle}{L2GTX: From Local to Global Time Series Explanations}

\hypersetup{
pdftitle={L2GTX: From Local to Global Time Series Explanations},
pdfsubject={Explainable Artificial Intelligence, Time Series Analysis},
pdfauthor={Ephrem Tibebe Mekonnen, Luca Longo, Lucas Rizzo, Pierpaolo Dondio},
pdfkeywords={Explainable AI, Time Series, Global Explanations, Model Interpretability},
}

\begin{document}
\maketitle
\vspace{-0.5em}

\begin{abstract}
Deep learning models achieve high accuracy in time series classification, yet understanding their class-level decision behaviour remains challenging. Time series explanations must account for temporal dependencies and capture patterns that recur across instances. However, existing approaches for extracting such explanations face three key limitations: \textbf{(i)} model-agnostic XAI methods developed for images and tabular data do not readily extend to time series due to temporally structured and non-independent observations; \textbf{(ii)} global explanation synthesis for time series remains underexplored; and \textbf{(iii)} the few global approaches that exist are largely model-specific, limiting architecture-neutral interpretability.
We propose \textbf{L2GTX}, a fully model-agnostic method that generates class-wise global explanations by aggregating local explanations from a selective, representative set of time series instances. L2GTX first extracts the top local clusters of parameterised temporal event primitives, such as increasing or decreasing trends and local extrema, together with their associated importance scores, from instance-level explanations generated using \textbf{LOMATCE} (LOcal Model-Agnostic Time Series Classification Explanations). These clusters are merged across instances to reduce redundancy, and an instance–cluster importance matrix is constructed to estimate global relevance. Under a user-defined instance selection budget, L2GTX selects representative instances that maximise coverage of influential clusters. The events from the selected instances that belong to the covered clusters are then aggregated into concise, class-wise global explanations by summarising event attributes per primitive type.
Experiments on six benchmark time series datasets show that L2GTX produces compact and interpretable global explanations while maintaining stable global faithfulness, measured by mean local surrogate fidelity ($R^2$), across different levels of explanation consolidation. These results demonstrate that L2GTX enables faithful local-to-global explanation synthesis for time series classifiers without relying on internal model details.
\end{abstract}

\keywords{Explainable AI \and
time-series explainability \and
global explanations \and
event-based explanations \and
local-to-global aggregation \and
model-agnostic interpretability \and
time series classification}

\section{Introduction}
\label{sec:intro}

Deep learning models can achieve strong predictive performance in time series classification and are increasingly deployed across a wide range of application domains, including finance \cite{kim2019financial}, sensor monitoring \cite{shu2019energy}, and safety-critical settings such as healthcare \cite{liu2022arrhythmia, strodthoff2020deep}. Despite their accuracy, understanding a model’s class-level decision behaviour, that is, the temporal patterns it consistently relies on across instances and classes, remains challenging. In practice, most time series models operate as opaque black boxes: given an input sequence, they produce a prediction without providing human-understandable insight into the temporal cues that drive that decision. This lack of transparency limits trust, complicates error analysis, and raises concerns in settings where reliability, accountability, and regulatory compliance are required.

To address these challenges, a substantial body of research has emerged in explainable artificial intelligence (XAI) \cite{theissler2022explainable,longo2023explainable, vilone2021classification}. Existing explanation methods can be categorised along several dimensions, one of which distinguishes between local explanations, which aim to explain individual predictions, and global explanations, which seek to characterise a model’s overall decision-making behaviour. While local explanations are valuable for instance-level inspection and debugging, global explanations are essential for understanding systematic patterns, biases, and failure modes learned by a model.

Although XAI has produced mature and widely adopted methods for static modalities such as image and tabular data, extending these approaches to time series models is non-trivial \cite{theissler2022explainable, schlegel2019towards, rojat2021explainable}. Time series are characterised by strong temporal dependencies, variable-length events, and shifting pattern locations across instances. As a result, explanations must preserve temporal ordering, event duration, and the recurrence of patterns across samples. These constraints complicate the direct adoption of existing XAI techniques, particularly when the goal is to synthesise interpretable explanations at a global, class-wise level.

Consequently, most explainability methods for time series classifiers focus on local explanations \cite{di2022explainable, rojat2021explainable}, identifying salient time steps or subsequences for individual predictions. In contrast, global explanation methods for time series remain comparatively sparse and typically tied to specific model architectures.
There is therefore a clear need for a model-agnostic approach that can synthesise class-wise global explanations from local temporal patterns observed across a selective and representative set of instances.

In this work, we propose L2GTX (Local-to-Global Time-series eXplanations), a model-agnostic local-to-global explanation method for deep time series classifiers. L2GTX first extracts instance-level explanations using LOMATCE \cite{11216415}, which represents model behaviour in terms of parameterised temporal event primitives \cite{kadous1999learning,mekonnen2022interpreting}. These local event clusters are consolidated across instances to reduce redundancy, and an instance–cluster importance matrix is constructed to estimate global relevance. Under a fixed explanation budget, L2GTX selects a diverse and representative subset of instances that jointly cover influential temporal patterns. The events associated with the selected clusters are then aggregated into concise, class-wise global explanations that summarise typical event timing, duration (for monotonic trends), magnitude (for local extrema), and variability.

This research contributes with a model-agnostic method for synthesising class-wise global explanations in time series classification by selecting representative instances and aggregating structured temporal events aligned with globally influential clusters.

The remainder of this article is organised as follows. Section~\ref{sec:related_work} reviews related work in explainable time series classification as well as conceptually related approaches. Section~\ref{sec:methodology} presents the proposed L2GTX method. Section~\ref{sec:exp_setup} describes the experimental setup, parameter settings, and evaluation protocol. Section~\ref{sec:results} reports the results and discusses the findings. Finally, Section~\ref{sec:conclusion} concludes the paper and outlines directions for future work.

\section{Related Work}
\label{sec:related_work}

Explainable Artificial Intelligence seeks to make the behaviour of complex machine learning models more transparent, with much of the literature focusing on post-hoc explanations that analyse trained models \cite{theissler2022explainable, rojat2021explainable}. These methods are commonly divided into local explanations, which explain individual predictions, and global explanations, which aim to summarise a model’s overall behaviour \cite{molnar2025, vilone2021classification}.
Most work on explainability for time-series classifiers has focused on local explanations that identify temporal regions or events contributing to a specific prediction.
Many of these works adapt explainability methods originally developed for image or tabular data to sequential settings by treating individual time steps or contiguous segments \cite{schlegel2019towards, schlegel2021ts}.
 These local explanation methods include both model-specific and model-agnostic approaches.
Several model-specific approaches extend Class Activation Mapping (CAM) \cite{zhou2016learning} or Layer-wise Relevance Propagation (LRP) \cite{bach2015pixel} to time-series models \cite{zhou2021salience, vielhaben2023explainable}. Although effective in analysing internal model behaviour, these methods often depend on architectural assumptions or latent activations and tend to produce explanations that are more suitable for developers than for end users \cite{schlegel2021time, jeyakumar2020can}.

Model-agnostic methods such as LIME \cite{ribeiro2016should} and SHAP \cite{lundberg2017unified} have also been adapted to time-series settings \cite{schlegel2019towards}. Many of these adaptations treat time steps as independent features, which ignores temporal dependencies. To address this, several extensions perturb longer segments or homogeneous regions \cite{guilleme2019agnostic, neves2021interpretable, sivill2022limesegment}, while others incorporate temporal or frequency representations, such as TF-LIME \cite{wang2025tf}. 

More recently, LOMATCE (LOcal Model-Agnostic Time-series Classification Explanations) \cite{11216415} was proposed as a model-agnostic local explanation method for time-series classifiers. LOMATCE represents explanations using parameterised event primitives, such as increasing or decreasing trends and local extrema, yielding event-based explanations that align with how humans typically interpret time series. Despite their effectiveness for instance-level analysis, local explanations alone do not capture how explanatory patterns recur or stabilise across a dataset or class. However, deriving global explanations for time-series models also introduces further challenges. Global explanations must capture patterns that are not only important for individual predictions, but also consistent and representative across instances and classes. In time-series data, this requires consolidating temporally localised events that may vary in timing, duration, or magnitude, while avoiding redundancy and maintaining fidelity to the underlying model. As a result, global explanation methods for time-series classifiers remain limited \cite{rojat2021explainable, di2022explainable, theissler2022explainable}. For example, existing approaches often rely on model-specific components, such as activations or filters, to extract global insights \cite{oviedo2019fast, siddiqui2019tsviz, cho2020interpretation}, thereby limiting their general applicability.

A small number of model-agnostic methods attempt to provide global explanations for time-series classifiers using rule-based abstractions or event primitives. In particular, the works described in \cite{mekonnen7global, mekonnen2023explaining} propose global explainers that extract parameterised event primitives and derive explanations through decision-tree surrogate models that approximate black-box predictions. Hence, these methods derive global explanations through surrogate rule extraction rather than by aggregating local surrogate explanations from representative instances. As a result, global fidelity is determined by the surrogate’s approximation quality, with no explicit control over instance-level explanation fidelity, coverage, or representativeness.


Related ideas have also been explored outside the time-series domain. SP-LIME \cite{ribeiro2016should} selects representative instances whose local explanations reflect model behaviour, but does not aggregate explanations or produce class-wise summaries. Other methods, such as GLocalX \cite{setzu2021glocalx}, aggregate local explanations to form global models for tabular data. Similarly, global aggregation methods in natural language processing derive class-wise explanations over discrete feature spaces \cite{van2019global}, but do not model temporal structure.

In contrast, we propose L2GTX, a model-agnostic method that connects local and global explainability for time-series classifiers. L2GTX aggregates local explanation events into interpretable, class-wise global summaries using importance\-aware clustering and budgeted selection of representative instances. Section~\ref{sec:methodology} presents the proposed L2GTX method in detail.

\section{The L2GTX Method}
 \label{sec:methodology}

\begin{figure}
    \centering
    \includegraphics[width=\linewidth]{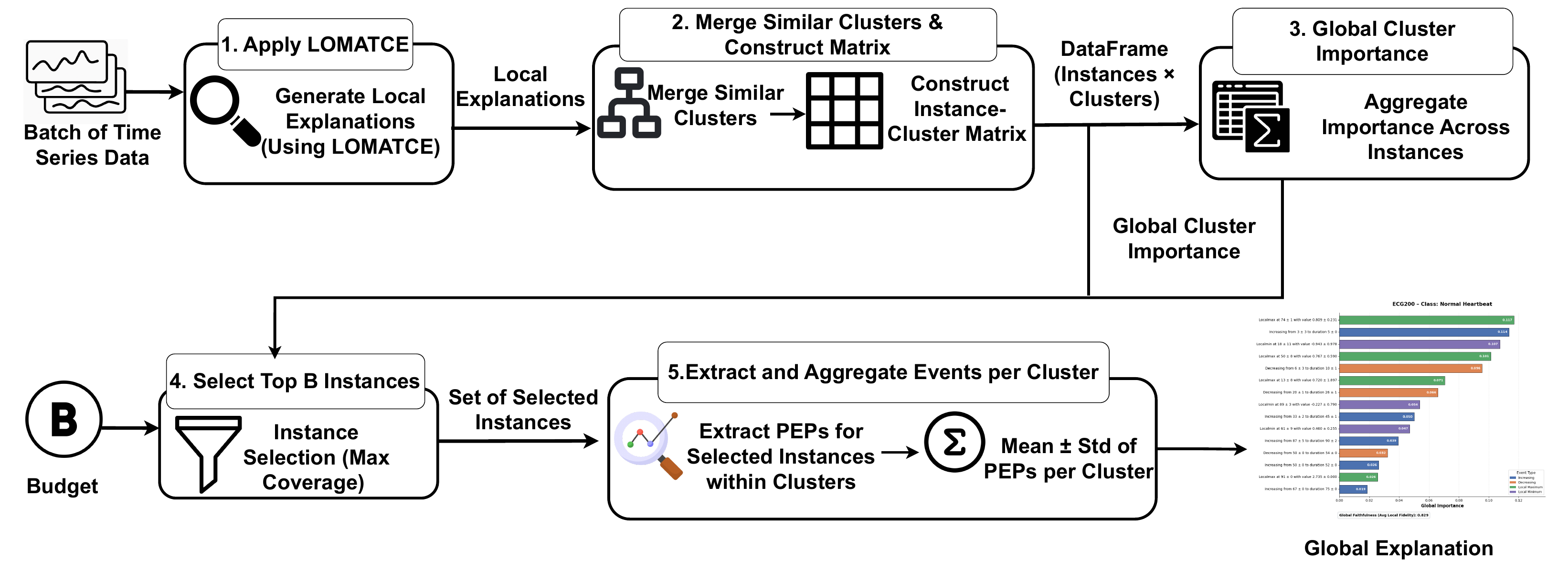}
    \caption{ Overview of the L2GTX method.
(1) Local explanations are generated for a batch of time-series instances using LOMATCE. (2) Similar event clusters are merged to reduce redundancy, producing an instance–cluster importance matrix. (3) Global importance scores are computed for the clusters. (4) Given an instance selection budget $B$, the instance–cluster importance matrix, and the global cluster importance scores, a representative subset of instances that maximises the coverage of influential clusters is selected. (5) Parameterised event primitives (PEPs) from the selected instances are aggregated to produce the final class-wise global explanation.}

    \label{fig:l2gtx}
\end{figure}

This section describes L2GTX (Local-to-Global Time series eXplanations), a method for deriving global insights from local, model-agnostic explanations of black-box time series classifiers.
Let the dataset be denoted as $\mathcal{X} = \{ (X_1, y_1), \dots, (X_m, y_m) \}$, where each $X_i \in \mathbb{R}^T$ is a univariate time series of length $T$ and $y_i \in \{1, \dots, C\}$ is its corresponding class label. As illustrated in Figure~\ref{fig:l2gtx}, the methodology comprises five sequential steps. 

To ensure class-balanced global explanations and control computational cost, L2GTX operates on a subset of $n_{\text{inst}}$ instances per class randomly sampled without replacement from $\mathcal{X}$, yielding a total of $N = C \cdot n_{\text{inst}}$ instances. All subsequent steps operate on this selected subset.

\subsection{Local Attribution via LOMATCE}

L2GTX begins by generating local explanations for a batch of time-series instances $\mathcal{X}$ using the LOMATCE algorithm~\cite{11216415}, which is detailed in Appendix~\ref{app:lomatce}. For each instance $X_i$, LOMATCE constructs a local neighbourhood of $S$ perturbed samples by randomly perturbing temporal segments and weighting each sample by its similarity to the original instance. From these samples, all Parameterised Event Primitives (PEPs) are extracted, including \emph{increasing} and \emph{decreasing} segments parameterised by $(\textit{start\_time}, \textit{duration}, \textit{avg\_gradient})$, and \emph{local maxima} and \emph{local minima} parameterised by $(\textit{time}, \textit{value})$. 

\begin{figure*}[t]
    \centering

    \begin{subfigure}{\linewidth}
        \centering
        \includegraphics[width=\linewidth]{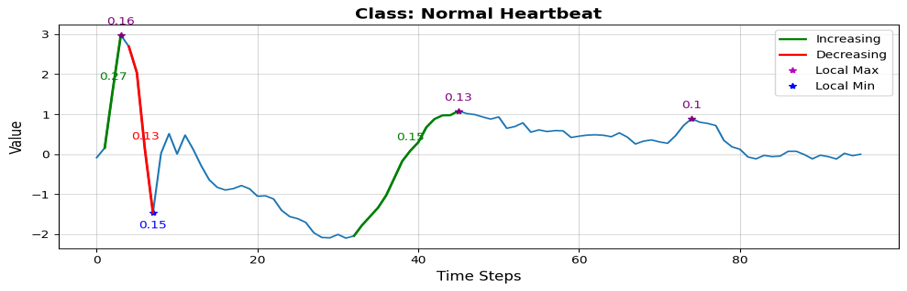}
        \caption{Normal heartbeat}
        \label{fig:lomatce_normal}
    \end{subfigure}

    \vspace{0.8em}

    \begin{subfigure}{\linewidth}
        \centering
        \includegraphics[width=\linewidth]{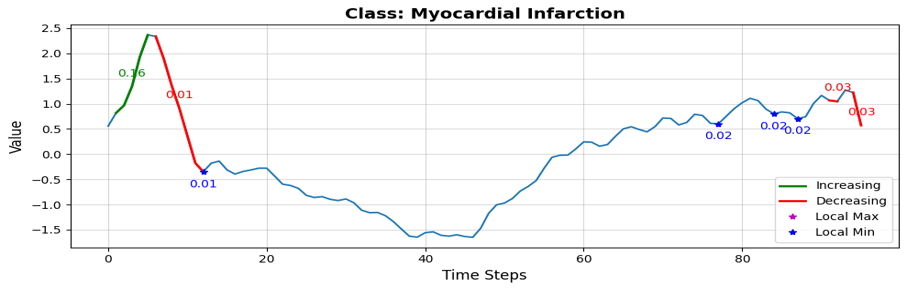}
        \caption{Myocardial infarction}
        \label{fig:lomatce_mi}
    \end{subfigure}

    \caption{Illustrative examples of local explanations generated by LOMATCE for the FCN classifier on the ECG200 dataset.}
    \label{fig:lomatce_ecg200}
\end{figure*}

For each PEP type, events from all neighbourhood samples are clustered independently using the K-means algorithm, with $K$ determined using the silhouette method. The number of events assigned to each cluster is organised into a structured event matrix $\mathbf{Z}_i \in \mathbb{R}^{S \times K}$, where rows correspond to perturbed samples, columns correspond to PEP clusters, and each entry records the event count for that sample–cluster pair. A weighted linear surrogate (ridge regression) is then trained on $\mathbf{Z}_i$ using the sample weights and the corresponding black-box predictions, yielding importance scores $\hat{\beta}_i \in \mathbb{R}^{K}$ for the clusters. L2GTX retains the top-$n$ PEP clusters per instance, analogous to selecting the top-$n$ features in feature-attribution methods \cite{ribeiro2016should}, based on the magnitude of their surrogate importance coefficients, producing a compact local explanation for subsequent aggregation. Figure~\ref{fig:lomatce_ecg200} shows illustrative examples of local explanations generated by LOMATCE for the FCN classifier on the ECG200 dataset.

\subsection{Merge Similar Clusters \& Construct Instance-Cluster Matrix}

For each instance $X_i$ in the dataset, the LOMATCE explainer produces a set of local clusters $C_{i,k}$, representing distinct subsequences or events (e.g., \textit{Increasing\_cluster\_1}, \textit{LocalMax\_cluster\_2}). Since clusters differ across instances, similar clusters of the same PEP type are merged to obtain a concise global representation.
We perform event-wise agglomerative hierarchical clustering based on the Euclidean distance between cluster centroids. Let $\mathbf{\mu}_{i,k}$ denote the centroid of cluster $C_{i,k}$. For each event type $e \in \{\text{Increasing, Decreasing, LocalMax, LocalMin}\}$, all centroids of type $e$ are clustered across instances. Let $\{d_r\}$ denote the set of merge distances produced by the hierarchical clustering procedure. Given a user-defined merge percentile $p$, we compute the cut distance $\tau = \mathrm{percentile}_p(\{d_r\})$ and obtain global clusters by cutting the dendrogram at distance $\tau$. Higher values of $p$ yield larger $\tau$ and therefore fewer, more consolidated global clusters, while lower values preserve more fine-grained distinctions.
Formally, the merged cluster mapping is
\begin{equation}
f_{\text{merge}} : C_{i,k} \mapsto G_j, \quad G_j \in \mathcal{G}_e,
\end{equation}
where $\mathcal{G}_e$ denotes the set of global clusters for event type $e$.

After merging, we construct the instance-cluster matrix $\mathbf{M} \in \mathbb{R}^{N \times |\mathcal{G}|}$, where $N$ is the number of instances and $|\mathcal{G}|$ is the total number of global clusters across all event types. Each entry represents the importance of a global cluster $G_j$, for instance $i$:
\begin{equation}
M_{i,j} = \sum_{C_{i,k} \in G_j} I(C_{i,k}),
\end{equation}
where $I(C_{i,k})$ is the importance score of the local cluster $C_{i,k}$ assigned by LOMATCE.

\subsection{Compute Global Cluster Importance}

To summarise the contribution of each global cluster across the dataset, we adopt the aggregation strategy from SP-LIME~\cite{ribeiro2016should}, adapting it to time-series clusters. 
Given the instance–cluster matrix $\mathbf{M} \in \mathbb{R}^{N \times G}$, the global importance of cluster $j$ is computed as:

\begin{equation}
I_j = \sqrt{\sum_{i=1}^{N} |M_{i,j}|}, \quad j = 1, \dots, G,
\end{equation}

where $N$ is the number of instances, $G$ is the total number of global clusters, and $M_{i,j}$ represents the importance of cluster $j$ for instance $i$.

Unlike SP-LIME, which aggregates feature importances for tabular models, here we aggregate cluster-level importance scores derived from LOMATCE, treating each global cluster of Parameterised Event Primitives (PEPs) as a “feature” summarising temporal patterns. The resulting vector $\mathbf{I} = [I_1, \dots, I_G]$ captures the overall significance of each global cluster and serves as the basis for selecting representative and non-redundant instances in the next step.

\subsection{Select Top B Instances}

Given the instance–cluster matrix $\mathbf{M} \in \mathbb{R}^{N \times |\mathcal{G}|}$, where $M_{i,j}$ represents the importance of global cluster $G_j$ for instance $i$, and the global cluster importance vector $\mathbf{I} \in \mathbb{R}^{|\mathcal{G}|}$, 
we select a subset of $B$ representative instances that collectively cover the most important clusters.

Let $S \subseteq \{1, \dots, N\}$ denote the set of selected instances, and let $\mathbf{c} \in \{0,1\}^{|\mathcal{G}|}$ be a coverage indicator vector, initially $\mathbf{c} = \mathbf{0}$. 
The selection proceeds greedily: at each step, we choose the instance that maximises the marginal coverage of uncovered clusters weighted by their global importance:

\begin{equation}
i^* = \arg\max_{i \notin S} \sum_{j=1}^{|\mathcal{G}|} I_j \cdot \mathbf{1}\{M_{i,j} > 0 \land c_j = 0\},
\end{equation}

where $\mathbf{1}\{\cdot\}$ is the indicator function. After selecting $i^*$, the coverage vector is updated to mark newly covered clusters:

\begin{equation}
c_j \gets \max(c_j, \mathbf{1}\{M_{i^*,j} > 0\}), \quad \forall j.
\end{equation}

This process is repeated until $|S| = B$, ensuring that the selected instances maximise coverage of the most globally important clusters while respecting the user-specified budget $B$.

\subsection{Extract \& Aggregate Events per Global Cluster}

After selecting the top $B$ representative instances, we identify the global clusters they cover. Each global cluster $G_j$ originally contains multiple local clusters $C_{i,k}$ from the selected instances, each with a set of events. To simplify the representation and produce a concise global explanation, we remove the local cluster level and aggregate all events from the selected instances directly into the corresponding global cluster:

\begin{equation}
\mathcal{E}_j = \bigcup_{i \in S} \bigcup_{k} \mathcal{E}_{i,k}.
\end{equation}
Here, the first union iterates over all selected instances, and the second union iterates over the local clusters within each instance. This flattening ensures that all events contributing to a global cluster are included. Importantly, since each global cluster inherently encodes the event type (e.g., Increasing, Decreasing, LocalMax, LocalMin), the local cluster structure is no longer needed for summarisation.
For each event attribute $a \in \{\text{start\_time}, \text{duration}, \text{time}, \text{value}\}$ 
of the events in $\mathcal{E}_j$, we compute the mean and standard deviation:
\begin{align}
\mu_j^{a} = \frac{1}{|\mathcal{E}_j|} \sum_{e \in \mathcal{E}_j} a(e)
\\
\sigma_j^{a} = \sqrt{\frac{1}{|\mathcal{E}_j|} \sum_{e \in \mathcal{E}_j} (a(e) - \mu_j^{a})^2 }.
\end{align}

These statistics are then used to summarise the global cluster by event type. For \emph{increasing} and \emph{decreasing} events, the cluster is described as starting at $\mu_j^{\text{start\_time}} \pm \sigma_j^{\text{start\_time}}$ and extending over $\mu_j^{\text{duration}} \pm \sigma_j^{\text{duration}}$, representing the typical temporal range. For \emph{local maxima} and \emph{local minima}, the cluster is summarised at time $\mu_j^{\text{time}} \pm \sigma_j^{\text{time}}$ with value $\mu_j^{\text{value}} \pm \sigma_j^{\text{value}}$, capturing the typical occurrence and magnitude of the extrema.

This flattening and summarisation provides a concise, interpretable representation of each global cluster, forming class-level explanations that can be visualised and inspected to understand typical model behaviour. 


The complete L2GTX procedure is formally summarised in Algorithm~\ref{alg:l2gtx}.

\begin{algorithm}[htbp]
\caption{L2GTX: Local-to-Global Time Series Explanations}
\label{alg:l2gtx}
\begin{algorithmic}[1]
\State \textbf{Input:} Dataset $\mathcal{X}$, Black-box $f$, Instances per class $n_{\text{inst}}$, Budget $B$, Threshold $\tau$
\State \textbf{Output:} Global Explanation Summary $\mathcal{S}$

\State \textbf{Instance Selection}
\State $\mathcal{X}' \gets$ Sample $n_{\text{inst}}$ instances per class from $\mathcal{X}$
\State $N \gets |\mathcal{X}'|$

\State \textbf{Step 1: Local Attribution}
\For{each $X_i \in \mathcal{X}'$}
    \State $\{\mathcal{C}_i, \hat{\beta}_i\} \gets \text{LOMATCE}(X_i, f)$
\EndFor

\State \textbf{Step 2: Meta-Clustering}
\State $\mathcal{G} \gets \text{AgglomerativeClustering}(\{\mu_{i,k}\}, \tau)$
\State Construct $\mathbf{M} \in \mathbb{R}^{N \times |\mathcal{G}|}$ where $M_{i,j} = \text{Importance}(G_j, X_i)$

\State \textbf{ Step 3: Global Importance}
\State $I_j \gets \sqrt{\sum_{i=1}^{N} |M_{i,j}|}$ for all $G_j \in \mathcal{G}$

\State \textbf{Step 4: Submodular Selection}
\State $S \gets \emptyset, \mathbf{c} \gets \mathbf{0}$
\While{$|S| < B$}
    \State $i^* \gets \arg\max_{i \notin S} \sum_{j} I_j \cdot \mathbf{1}\{M_{i,j} > 0 \land c_j = 0\}$
    \State $S \gets S \cup \{i^*\}$; update $\mathbf{c}$
\EndWhile

\State \textbf{Step 5: Global Synthesis}
\State $\mathcal{S} \gets \text{AggregatePEPs}(\{G_j : j \in S\})$
\State \Return $\mathcal{S}$
\end{algorithmic}
\end{algorithm}

\section{Experimental Setup}
\label{sec:exp_setup}

This section details the empirical setup used to evaluate L2GTX, including the selection of datasets, the architectural specifications of the black-box models, and the quantitative evaluation metric used to assess explanation fidelity, with the expectation that local-to-global aggregation preserves surrogate fidelity while producing compact global summaries.

\subsection{Datasets}
We evaluate our method on six widely used univariate time series datasets from the UCR Archive \cite{dau2019ucr}. This selection comprises five real-world datasets (\textit{ECG200}, \textit{GunPoint}, \textit{Coffee}, \textit{FordA}, and \textit{FordB}) and one synthetic dataset (\textit{CBF}).
ECG200 contains electrocardiogram (ECG) signals representing individual heartbeats, with the task of distinguishing normal heartbeats from those associated with myocardial infarction. GunPoint consists of hand movement trajectories performed by two actors, where the goal is to classify the gestures as either gun-draw or point. The Coffee dataset contains spectrographic measurements of coffee beans, with two classes, Arabica and Robusta. FordA and FordB contain time series data of engine noise collected during standard operating conditions, where the task is to classify the presence or absence of fault symptoms. However, the FordB dataset is gathered in a noisier environment. Finally, CBF is a synthetic dataset composed of three characteristic patterns: cylinder, bell, and funnel. The descriptive statistics for these datasets are summarised in Table \ref{tab:dataset}.

\begin{table}[hbtp]
    \centering
    \small
    \setlength{\tabcolsep}{10pt} 
    \caption{Descriptive statistics of the six UCR datasets used in the experiment.}
    \begin{tabular}{lcccc}
    \toprule
    \textbf{Name} & \textbf{Data Size} & \textbf{No. Classes} & \textbf{Length} & \textbf{Type} \\
    \midrule
    ECG200   & 200  & 2 & 96  & Medical \\
    GunPoint & 200  & 2 & 150 & Action \\
    Coffee   & 48   & 2 & 286 & Food \\
    FordA    & 4921 & 2 & 500 & Sensor \\
    FordB    & 4446 & 2 & 500 & Sensor \\
    CBF      & 900  & 3 & 128 & Synthetic \\
    \bottomrule
    \end{tabular}
    \label{tab:dataset}
\end{table}


\subsection{Black-box Classifiers and Training}

The datasets were partitioned into training (70\%), validation (15\%), and test (15\%) sets. Minimal preprocessing was applied, restricted to batch-wise standardisation using the \texttt{TSStandardize()} function from the \texttt{tsai} library \cite{tsai} to ensure zero mean and unit variance.

The model-agnostic nature of L2GTX is demonstrated by its applicability to two distinct deep learning architectures. The first is a Fully Convolutional Network (FCN) comprising three 1D convolutional layers with filter sizes (128, 256, 128) and kernel sizes  sizes of (7, 5, 3).  Each layer is followed by batch normalisation and a ReLU activation, culminating in a Global Average Pooling (GAP) layer and a softmax output \cite{wang2017time}. The second architecture is an LSTM-FCN that incorporates a parallel Long Short-Term Memory (LSTM)  to capture long-range temporal dependencies. 

Both models were trained with early stopping, with patience set to 15 epochs to prevent overfitting. In addition, to ensure accuracy and stability, the model was trained 100 times with randomised splits for training, validation, and testing. The resulting validation and test accuracies are reported in Table~\ref{tab:results}. 

\begin{table}[htbp]
\centering
\caption{Mean test and validation accuracy (Test Acc and Valid Acc) for each dataset--model pairs, reported as mean ± 95\% confidence
interval.}
\label{tab:results}
\setlength{\tabcolsep}{12pt}
\renewcommand{\arraystretch}{1.2}
\begin{tabular}{lcccc}
\toprule
\textbf{Dataset} & \multicolumn{2}{c}{\textbf{FCN}} & \multicolumn{2}{c}{\textbf{LSTM-FCN}} \\
\cmidrule(lr){2-3} \cmidrule(lr){4-5}
& \textbf{Test} & \textbf{Val} & \textbf{Test} & \textbf{Val} \\
\midrule
ECG200   & 0.87 $\pm$ 0.03 & 0.86 $\pm$ 0.04 & 0.86 $\pm$ 0.01 & 0.85 $\pm$ 0.01 \\
GunPoint & 0.99 $\pm$ 0.01 & 0.98 $\pm$ 0.02 & 0.98 $\pm$ 0.01 & 0.98 $\pm$ 0.01 \\
Coffee   & 1.00 $\pm$ 0.00 & 1.00 $\pm$ 0.00 & 1.00 $\pm$ 0.00 & 1.00 $\pm$ 0.00 \\
FordA    & 0.90 $\pm$ 0.02 & 0.90 $\pm$ 0.02 & 0.88 $\pm$ 0.01 & 0.87 $\pm$ 0.01 \\
FordB    & 0.88 $\pm$ 0.02 & 0.89 $\pm$ 0.02 & 0.86 $\pm$ 0.01 & 0.86 $\pm$ 0.01 \\
CBF      & 0.98 $\pm$ 0.01 & 0.97 $\pm$ 0.01 & 1.00 $\pm$ 0.00 & 0.99 $\pm$ 0.01 \\
\bottomrule
\end{tabular}
\end{table}

\subsection{L2GTX Configuration}

For L2GTX global explanation synthesis, we sampled $n_{\text{inst}} = 15$ instances per class for small and medium-sized datasets, and $n_{\text{inst}} = 30$ for larger datasets (FordA, FordB, and CBF) to ensure adequate class coverage while maintaining computational feasibility. The instance selection budget $B$ specifies the maximum number of instances chosen by submodular optimisation and was set equal to $n_{\text{inst}}$.
 These sampled instances are used solely for explanation synthesis and do not influence the training or evaluation of the underlying classifiers.

Multiple agglomerative merge percentiles ($p \in \{25, 50, 75, 95\}$) were evaluated to assess robustness, as reported in the metric results. For visualisation of global explanations, $p = 95$ was adopted as the default setting, as it yields compact yet behaviour-preserving cluster summaries while retaining the dominant temporal events.

Local explanations were generated using the LOMATCE explainer with its default parameter settings as described in~\cite{11216415}, including neighbourhood sampling, event clustering, and silhouette-based estimation of the number of clusters.

\subsection{Global Faithfulness and Stability}

The primary evaluation metric is \textbf{Global Faithfulness (GF)}, which quantifies how accurately a global explanation summary reflects the behaviour of the underlying black-box model. 
Given a summary set $\mathcal{S}$ with budget $B$ (i.e., $|\mathcal{S}| = B$), GF is defined as the mean \emph{local fidelity}, measured using the coefficient of determination ($R^2$), across the selected instances:
\begin{equation}
    \mathrm{GF}(\mathcal{S}) = \frac{1}{|\mathcal{S}|} \sum_{x_i \in \mathcal{S}} F(x_i),
\end{equation}
where $F(x_i)$ denotes the local surrogate fidelity for instance $x_i$.

To assess the stability and reproducibility of the explanations, all experiments were conducted using three independent random seeds. 
We report the macro-averaged GF across classes together with the corresponding 95\% confidence intervals (CI).

\section{Results and Discussion}
\label{sec:results}


The quantitative results in Tables~\ref{tab:fcn_faithfulness} and~\ref{tab:faithfulness_lstmfcn} show that global faithfulness (GF) remains robust under varying degrees of consolidation. The consistent GF values observed for merge percentiles ($p \in {25, 50, 75, 95}$), together with overlapping confidence intervals for nearly all model–dataset pairs, indicate that L2GTX effectively compresses the explanation space into a compact set of global clusters without sacrificing global faithfulness to the underlying model behaviour.

\begin{table}[htbp]
\centering
\setlength{\tabcolsep}{5.8pt} 
\caption{Global faithfulness (mean $\pm$ 95\% CI) of L2GTX explanations using the FCN classifier under different merge percentiles. Results are macro-averaged across classes and aggregated over three random seeds.}
\label{tab:fcn_faithfulness}
\begin{tabular}{lcccc}
\toprule
\textbf{Dataset} & \textbf{p = 25} & \textbf{p = 50} & \textbf{p = 75} & \textbf{p = 95} \\
\midrule
ECG200   & 0.784 $\pm$ 0.015 & 0.788 $\pm$ 0.013 & 0.780 $\pm$ 0.026 &  0.792 $\pm$ 0.014 \\
GunPoint & 0.593 $\pm$ 0.007 & 0.599 $\pm$ 0.019 & 0.601 $\pm$ 0.007 & 0.597 $\pm$ 0.011 \\
Coffee   &  0.683 $\pm$ 0.010 & 0.678 $\pm$ 0.006 & 0.678 $\pm$ 0.005 & 0.678 $\pm$ 0.015 \\
FordA    & 0.674 $\pm$ 0.021 & 0.672 $\pm$ 0.029 & 0.673 $\pm$ 0.021 & 0.672 $\pm$ 0.028 \\
FordB    & 0.675 $\pm$ 0.008 & 0.679 $\pm$ 0.034 & 0.673 $\pm$ 0.006 & 0.673 $\pm$ 0.029 \\
CBF      & 0.625 $\pm$ 0.018 & 0.626 $\pm$ 0.011 & 0.633 $\pm$ 0.016 & 0.625 $\pm$ 0.008 \\
\bottomrule
\end{tabular}
\end{table}

\begin{table}[htbp]
\centering
\setlength{\tabcolsep}{5.9pt} 
\caption{Global faithfulness (mean $\pm$ 95\% CI) of L2GTX explanations using the LSTM-FCN classifier under different merge percentiles. Results are macro-averaged across classes and aggregated over three random seeds.}
\label{tab:faithfulness_lstmfcn}
\begin{tabular}{lcccc}
\toprule
\textbf{Dataset} & \textbf{p = 25} & \textbf{p = 50} & \textbf{p = 75} & \textbf{p = 95} \\
\midrule
ECG200     & 0.828 $\pm$ 0.010 & 0.832 $\pm$ 0.013 & 0.829 $\pm$ 0.021 & 0.831 $\pm$ 0.007 \\
GunPoint   & 0.617 $\pm$ 0.074 & 0.619 $\pm$ 0.067 & 0.588 $\pm$ 0.086 & 0.638 $\pm$ 0.011 \\
Coffee     & 0.617 $\pm$ 0.008 & 0.609 $\pm$ 0.004 & 0.616 $\pm$ 0.036 & 0.608 $\pm$ 0.003 \\
FordA      & 0.618 $\pm$ 0.028 & 0.621 $\pm$ 0.015 & 0.614 $\pm$ 0.039 & 0.627 $\pm$ 0.035 \\
FordB      & 0.661 $\pm$ 0.021 & 0.656 $\pm$ 0.039 & 0.651 $\pm$ 0.050 & 0.655 $\pm$ 0.027 \\
CBF        & 0.519 $\pm$ 0.020 & 0.508 $\pm$ 0.025 & 0.519 $\pm$ 0.033 & 0.502 $\pm$ 0.015 \\
\bottomrule
\end{tabular}
\end{table}

\begin{figure}[htbp]
    \centering

    \begin{subfigure}{\linewidth}
        \centering
        \includegraphics[width=\linewidth, ,height=0.40\textheight,keepaspectratio]{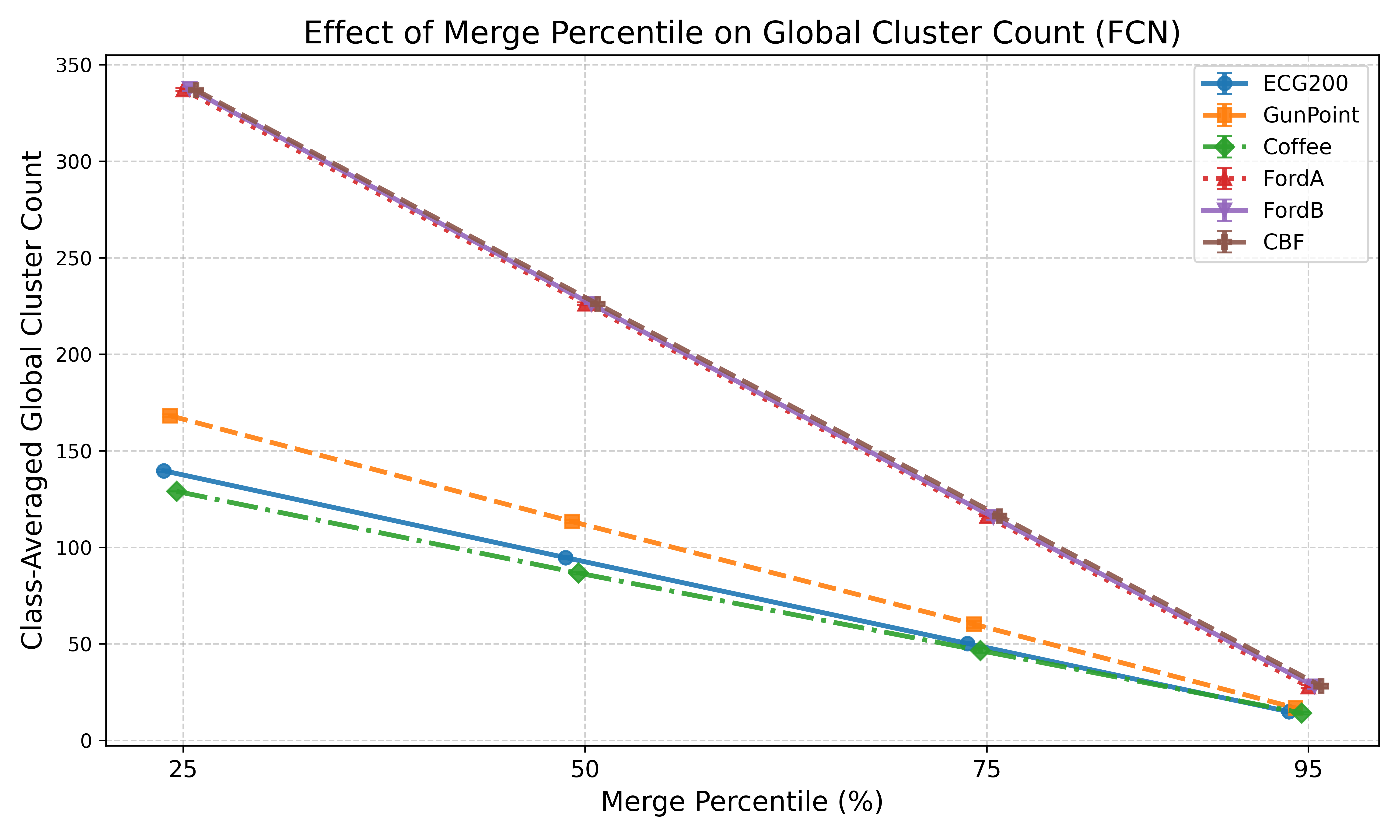}
        \caption{FCN}
        \label{fig:fcn_macro_clusters}
    \end{subfigure}

    \vspace{0.8em}

    \begin{subfigure}{\linewidth}
        \centering
        \includegraphics[width=\linewidth, ,height=0.40\textheight,keepaspectratio]{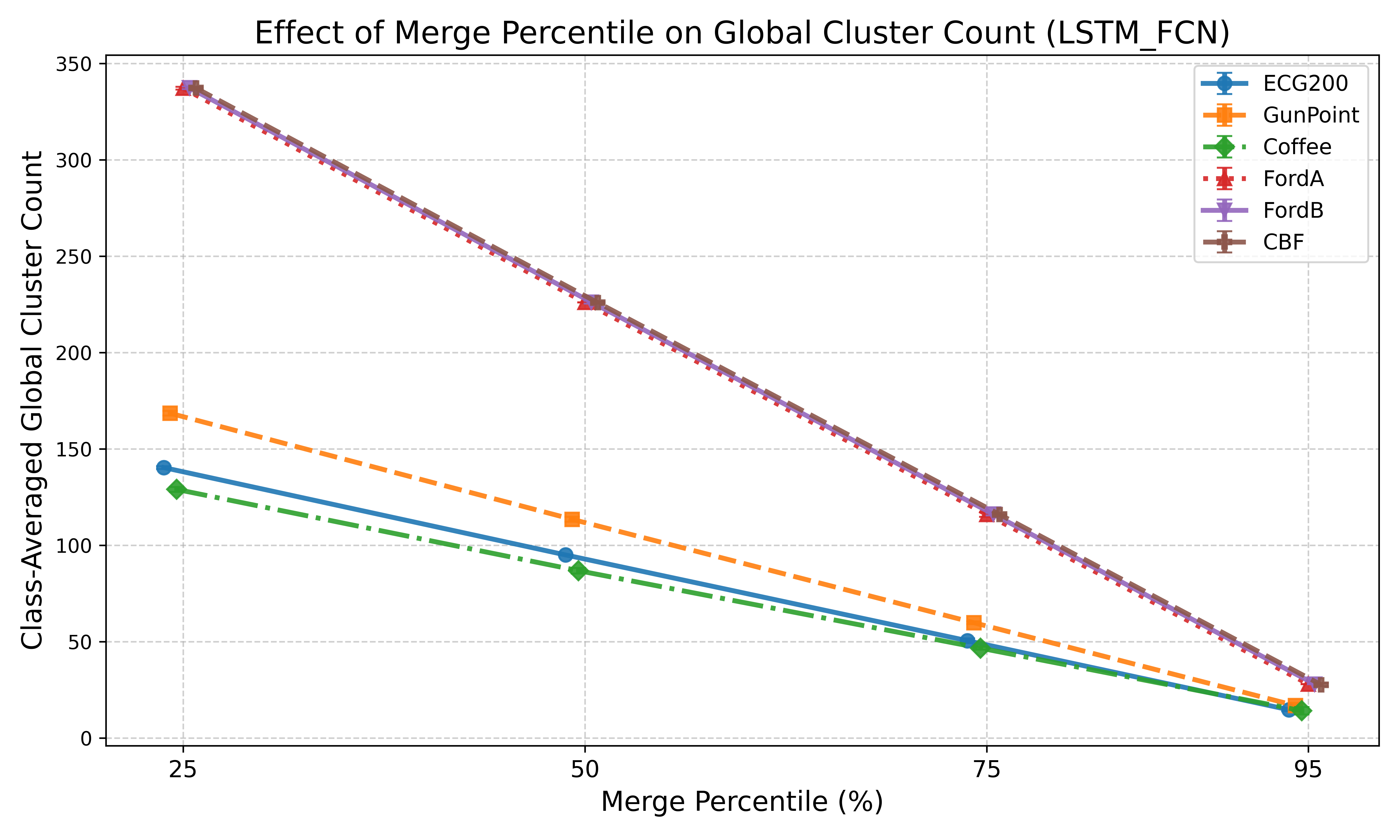}
        \caption{LSTM-FCN}
        \label{fig:lstmfcn_macro_clusters}
    \end{subfigure}

    \caption{
    Effect of the agglomerative clustering threshold (merge percentile) on the number of class-averaged global clusters.
    Increasing the merge percentile progressively consolidates event clusters, resulting in a monotonic reduction in the total number of global clusters.
    Across both models, this consolidation does not degrade explanation faithfulness, as reflected by stable or improved class-averaged global faithfulness scores.
    }
    \label{fig:macro_clusters_vs_merge}
\end{figure}

Figure~\ref{fig:macro_clusters_vs_merge} further illustrates this structural effect: as $p$ increases, the number of global clusters decreases monotonically, producing a more compact explanation space. In practice, we present results for $p=95$, which provides a compact representation while still capturing the dominant temporal event clusters in the resulting global explanations. This indicates that L2GTX retains shared decision-relevant cues rather than dispersing or discarding them during consolidation.



Most existing XAI methods applied to time-series data are adapted to provide local explanations at the level of individual time steps or predefined temporal segments. While such representations can indicate where a model assigns importance, they often offer limited interpretability: stating that a particular segment or time step is important does not explain what temporal behaviour within that region drives the decision. In contrast, describing an increasing trend over a temporal interval or identifying a pronounced local maximum provides a semantically meaningful explanation that aligns with how humans reason about time series signals. Aggregating importance values alone therefore reduces the complex temporal structure to a coarse ``bag of events'' that lacks morphological specificity. By explicitly parameterising explanations in terms of structural behavioural primitives, L2GTX provides insight into the ``why'' behind the ``where'' of a model’s attention. For example, identifying a top-ranked local maximum or increasing trend conveys a recognisable signal behaviour, whereas ranking a specific time step or segment primarily reflects position rather than temporal pattern. Such behavioural primitives can also be naturally translated into domain-specific semantics; for instance, an increasing trend may correspond to a spike in physiological signals in healthcare or a bullish movement in financial time series.

\begin{figure}[p]
    \centering

    \begin{subfigure}{\textwidth}
        \centering
        \includegraphics[width=\textwidth ,height=0.40\textheight,keepaspectratio]{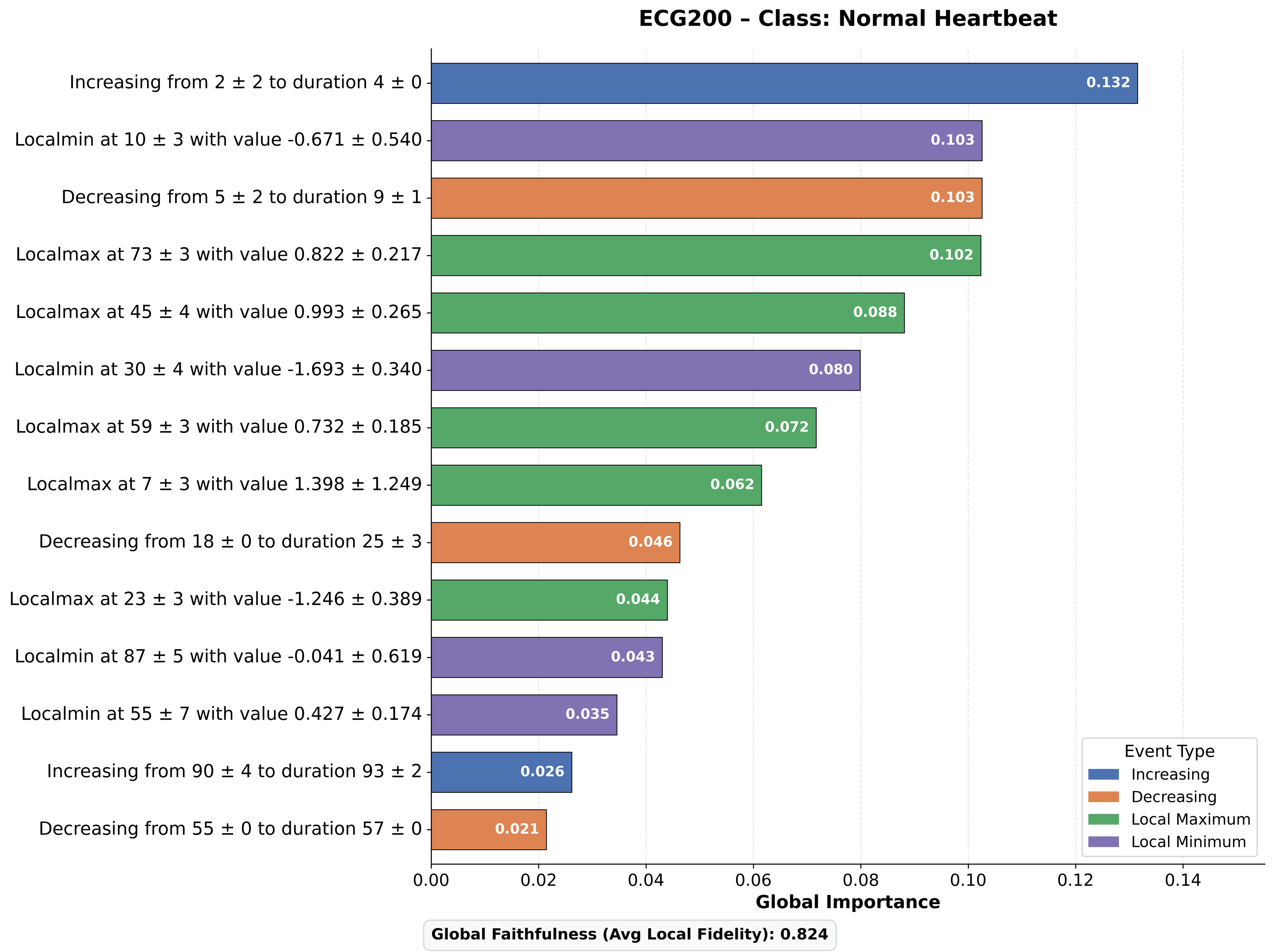}
        \caption{FCN – Normal}
        \label{fig:ecg200_fcn_normal}
    \end{subfigure}

    \vspace{0.6em}

    \begin{subfigure}{\textwidth}
        \centering
        \includegraphics[width=\textwidth,height=0.40\textheight,keepaspectratio ]{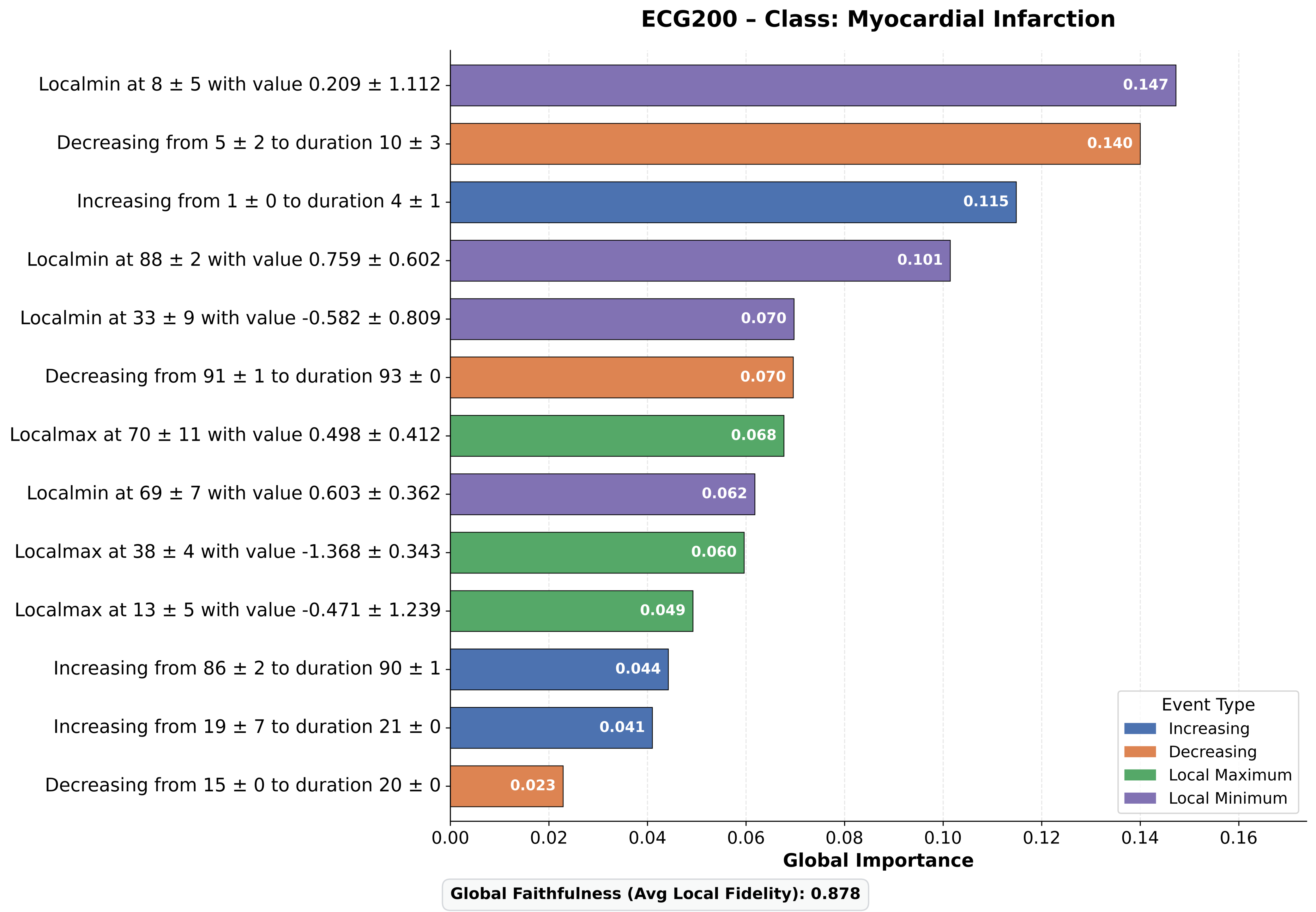}
        \caption{FCN – Infarction}
        \label{fig:ecg200_fcn_infarction}
    \end{subfigure}

    \caption{
    Class-wise global explanations produced by L2GTX for the ECG200 dataset using the FCN model.
    Bars indicate the global importance of aggregated temporal event clusters, with colours denoting event types.
    }
    \label{fig:ecg200_fcn_global}
\end{figure}

\begin{figure}[!htbp]
    \centering

    \begin{subfigure}{\textwidth}
        \centering
        \includegraphics[width=\textwidth,height=0.40\textheight,keepaspectratio]{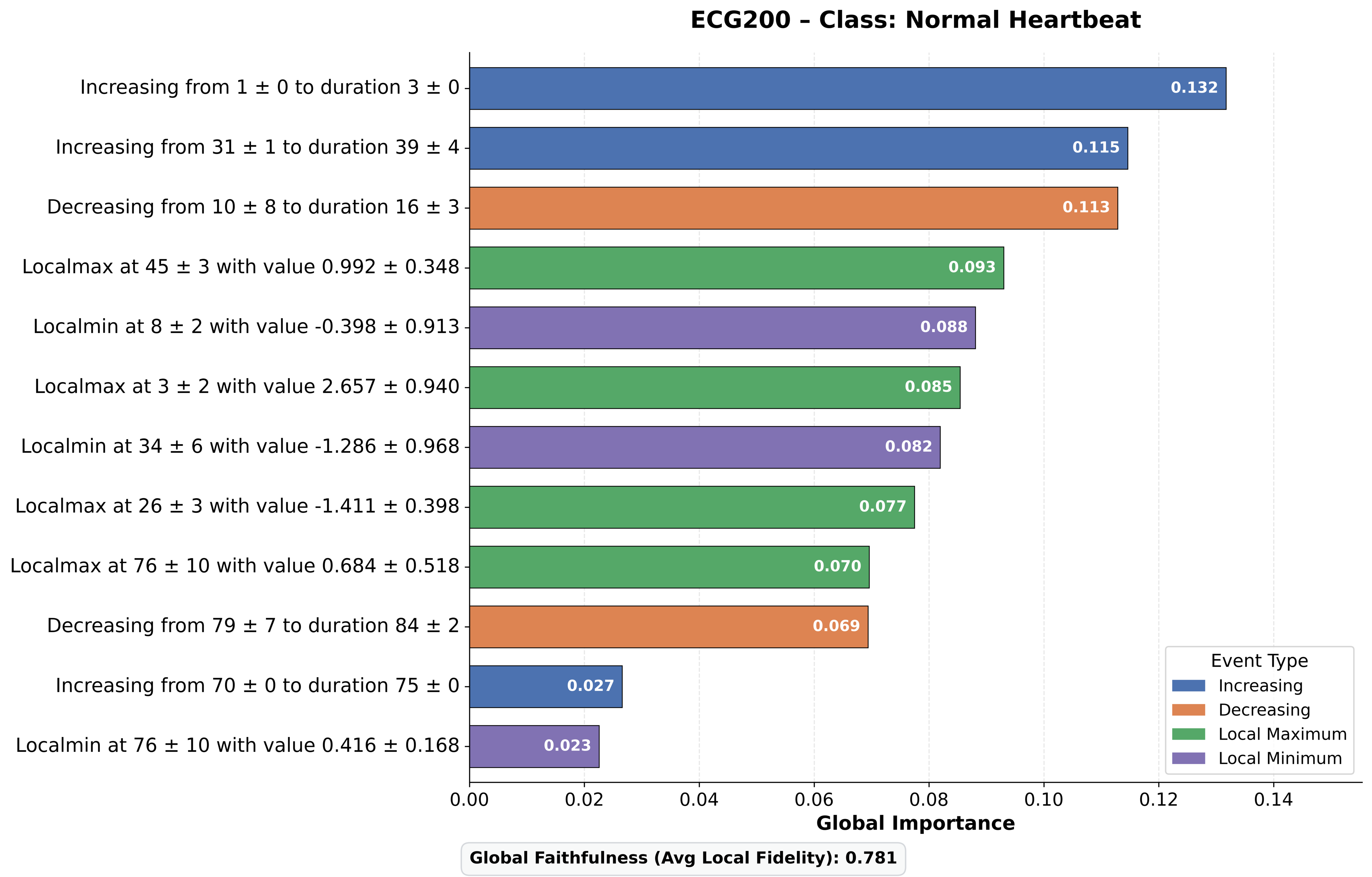}
        \caption{LSTM-FCN – Normal}
        \label{fig:ecg200_lstm-fcn_normal}
    \end{subfigure}

    \vspace{0.6em}

    \begin{subfigure}{\textwidth}
        \centering
        \includegraphics[width=\textwidth ,height=0.40\textheight,keepaspectratio]{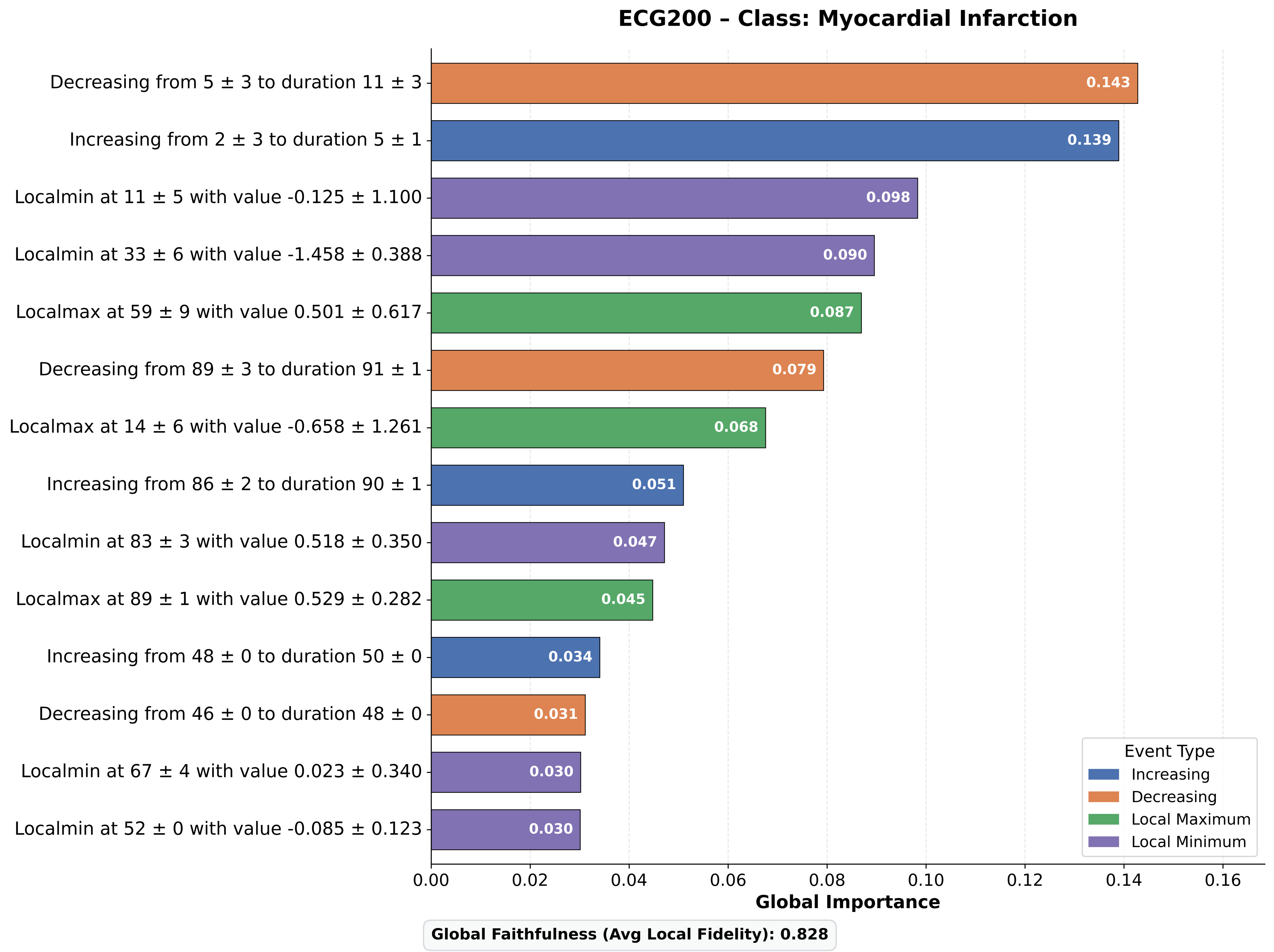}
        \caption{LSTM-FCN – Infarction}
        \label{fig:ecg200_lstm-fcn_infarction}
    \end{subfigure}

    \caption{
    Class-wise global explanations produced by L2GTX for the ECG200 dataset using the LSTM-FCN model.
    Bars indicate the global importance of aggregated temporal event clusters, with colours denoting event types.
    }
    \label{fig:ecg200_lstmfcn_global}
\end{figure}

\begin{figure}[!htbp]
    \centering

    \begin{subfigure}{\textwidth}
        \centering
        \includegraphics[width=\textwidth,height=0.40\textheight,keepaspectratio]{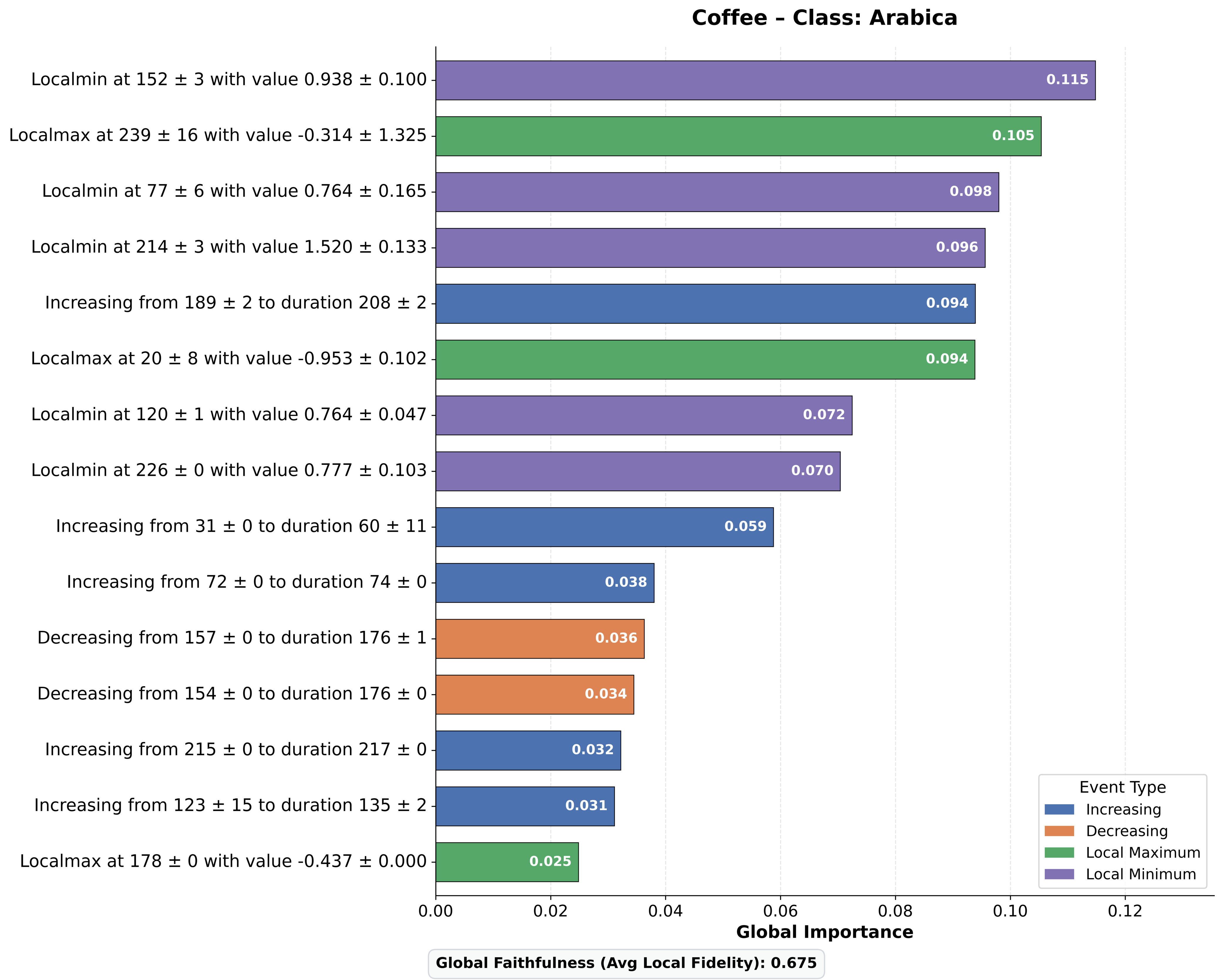}
        \caption{FCN – Arabica}
        \label{fig:coffee_fcn_arabica}
    \end{subfigure}

    \vspace{0.6em}

    \begin{subfigure}{\textwidth}
        \centering
        \includegraphics[width=\textwidth ,height=0.40\textheight,keepaspectratio]{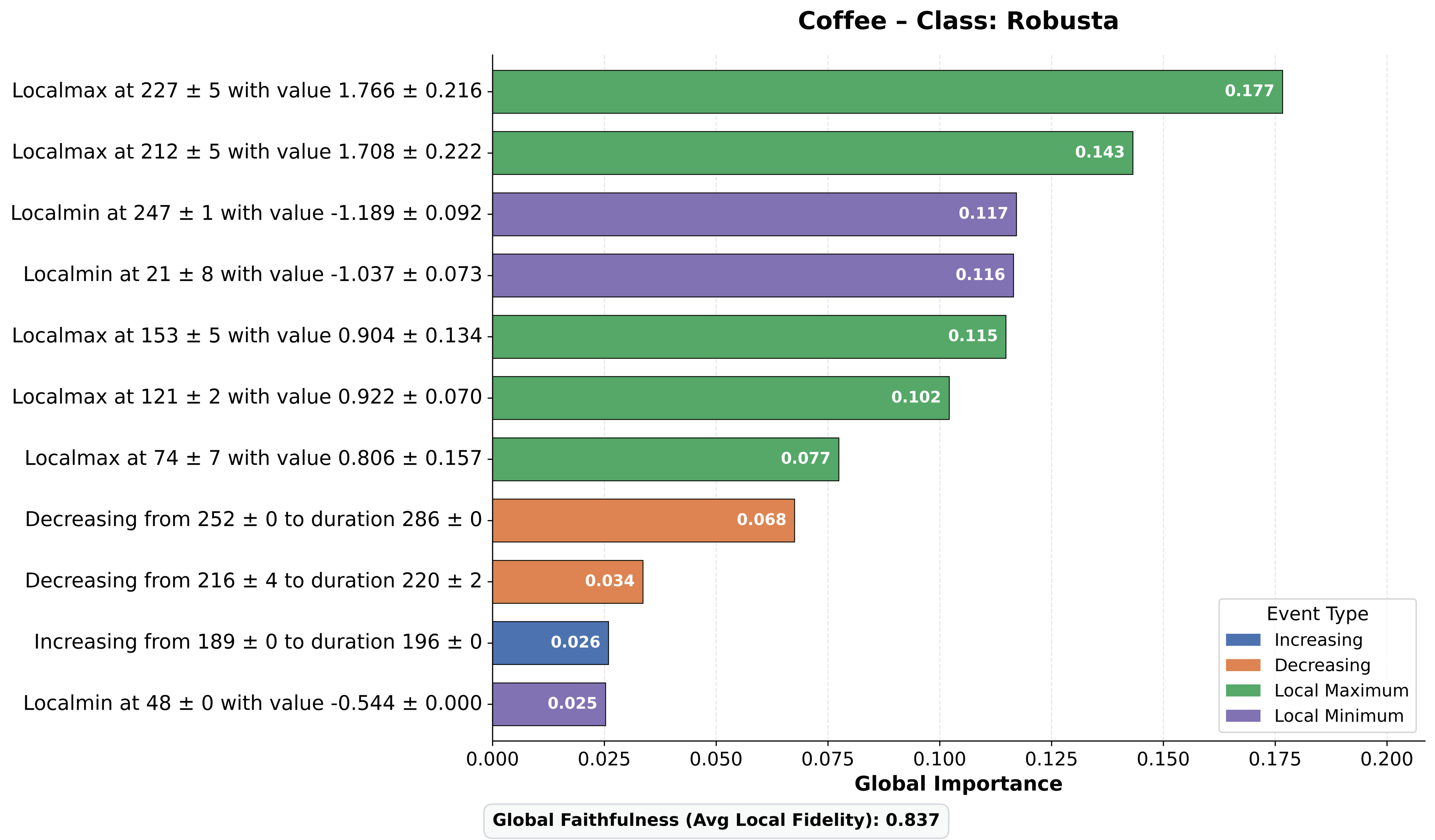}
        \caption{FCN – Robusta}
        \label{fig:coffee_fcn_robusta}
    \end{subfigure}

    \caption{
    Class-wise global explanations produced by L2GTX for the Coffee dataset using the FCN model at merge percentile 95.
    Bars denote the global importance of aggregated temporal event clusters, with colours indicating event dynamics.
    }
    \label{fig:coffee_fcn_global}
\end{figure}

\begin{figure}[!htbp]
    \centering

    \begin{subfigure}{\textwidth}
        \centering
        \includegraphics[width=\textwidth,height=0.45\textheight,keepaspectratio]{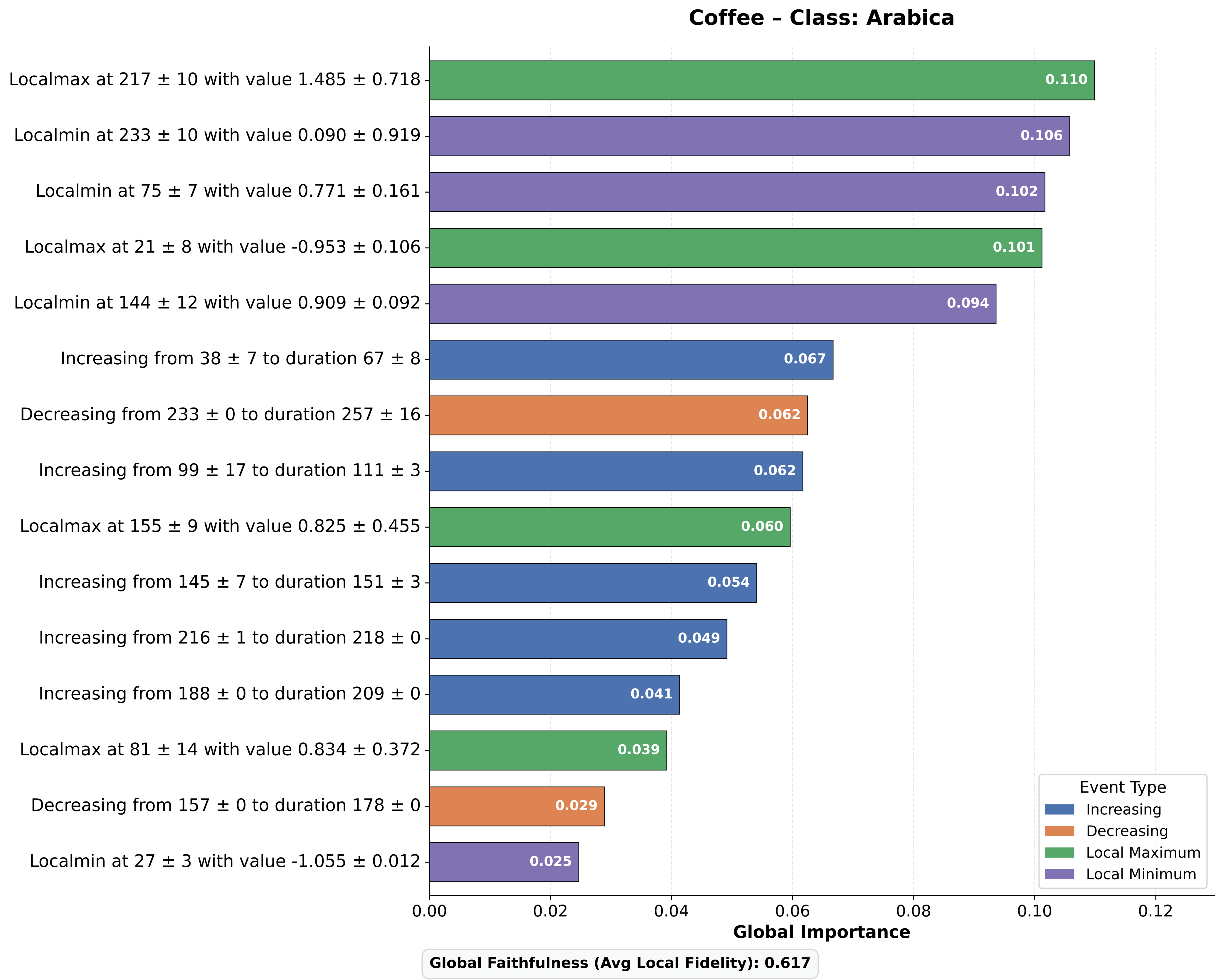}
        \caption{LSTM-FCN – Arabica}
        \label{fig:coffee_lstmfcn_arabica}
    \end{subfigure}

    \vspace{0.6em}

    \begin{subfigure}{\textwidth}
        \centering
        \includegraphics[width=\textwidth,height=0.45\textheight,keepaspectratio]{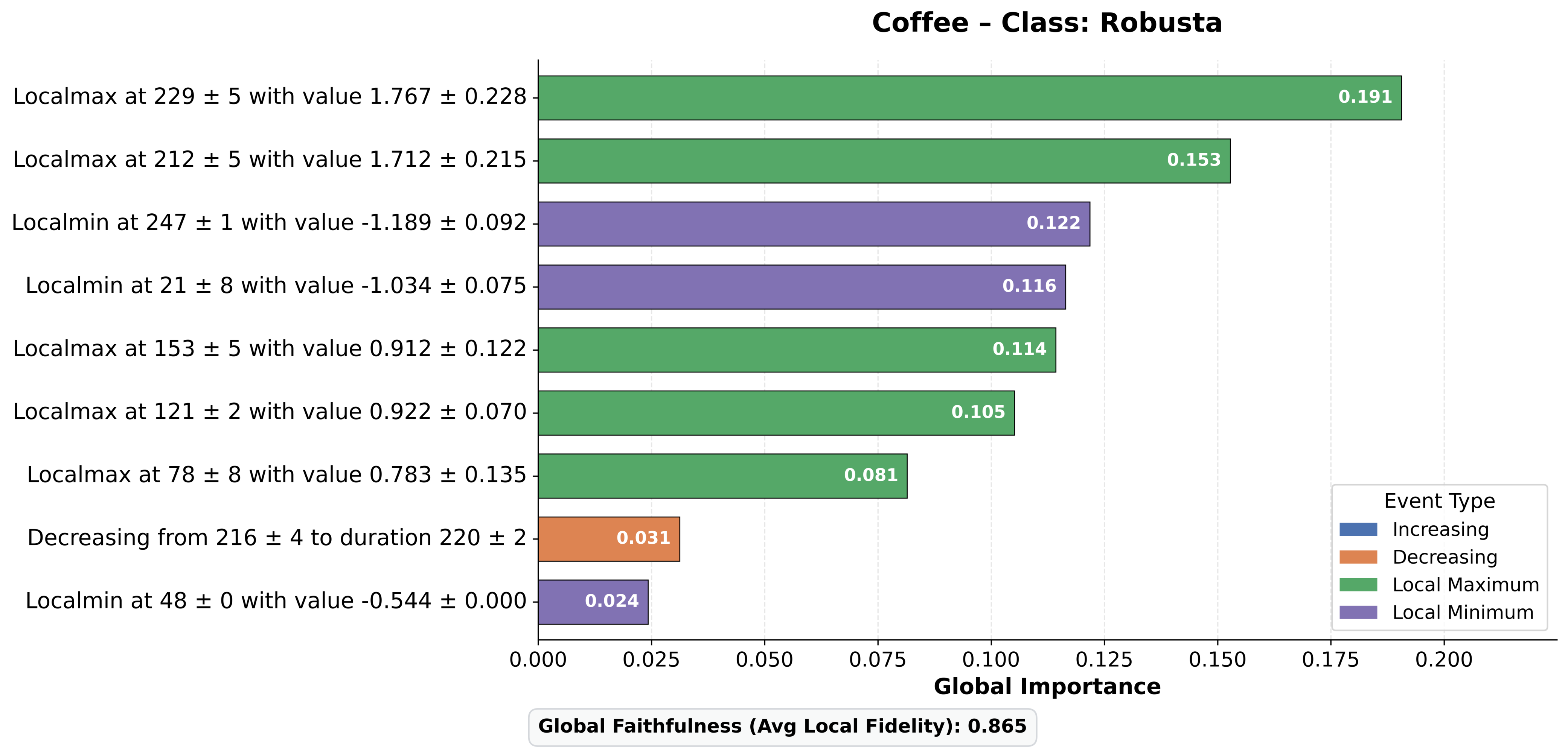}
        \caption{LSTM-FCN – Robusta}
        \label{fig:coffee_lstmfcn_robusta}
    \end{subfigure}

    \caption{
    Class-wise global explanations produced by L2GTX for the Coffee dataset using the LSTM-FCN model at merge percentile 95.
    }
    \label{fig:coffee_lstmfcn_global}
\end{figure}

We illustrate the interpretability of the resulting global explanations using two representative case studies, \textit{ECG200} and \textit{Coffee}, while quantitative faithfulness results are reported for all datasets. As shown in Figures~\ref{fig:ecg200_fcn_global}, \ref{fig:ecg200_lstmfcn_global}, \ref{fig:coffee_fcn_global}, and \ref{fig:coffee_lstmfcn_global}, L2GTX aggregates local events into concise global clusters and summarises them using attribute statistics.  The importance scores shown on the bars are normalised so that their sum equals one, representing the relative contribution of each event cluster to the global explanation for the class. Trend-based clusters are described by their temporal span, while extrema-based clusters are summarised using time--value statistics. This representation yields human-readable explanations that indicate where salient temporal events occur, how they manifest, and how consistently they appear across instances.

For example, previous work on distinguishing between Arabica and Robusta coffee beans has shown that Robusta samples tend to exhibit stronger overall spectral intensities, while Arabica samples are characterised by deeper absorption valleys and comparatively less dominant peaks~\cite{el2011discrimination}. As illustrated in Figures~\ref{fig:coffee_fcn_global} and~\ref{fig:coffee_lstmfcn_global}, both FCN and LSTM-FCN assign importance to overlapping spectral regions in the interval $t \in [210, 240]$, yet the resulting explanations differ in the types of temporal events emphasised. For the Robusta class, importance is concentrated on high-magnitude local maxima, whereas the Arabica class exhibits a more distributed pattern involving lower-intensity peaks alongside local minima. These differences are expressed through distinct configurations of extrema and accompanying spectral trends, consistent with how spectroscopic distinctions between coffee varieties are commonly discussed in the coffee spectroscopy literature.

Similarly, in the \textit{ECG200} dataset (Figures~\ref{fig:ecg200_fcn_global} and~\ref{fig:ecg200_lstmfcn_global}), the global explanations highlight distinct configurations of extrema and short temporal trends for the Normal and Myocardial Infarction classes. Normal heartbeats are characterised by a more distributed set of moderate-importance events, whereas infarction signals are dominated by fewer, more pronounced deflections and transitions. These patterns align with qualitative descriptions of ECG morphology commonly used to distinguish healthy and pathological cardiac signals, without relying on explicit clinical annotations.

Across both case studies, the global explanations consistently identify class-specific temporal regions composed of extrema and short directional trends. While the relative importance of individual clusters varies between FCN and LSTM-FCN, reflecting differences in architectural inductive biases, the explanations emphasise overlapping temporal regions for each class. This indicates that L2GTX captures behaviour-consistent global explanations while remaining sensitive to how different models rely on temporal information.

A practical limitation of the proposed method arises from the clustering of parameterised event primitives within LOMATCE, which constitutes the most computationally demanding stage of the explanation pipeline. This cost becomes more pronounced for long time series or when many neighbourhood samples are generated to construct local explanations. To mitigate this overhead, event clustering is parallelised on multicore architectures, and the number of clusters is estimated using a silhouette-based heuristic applied to a subset of the extracted events. While this strategy provides a favourable balance between computational efficiency and clustering stability, there remains room for further improvement in the scalability and efficiency of event-level clustering, particularly for large-scale or resource-constrained settings.

Overall, these results demonstrate that L2GTX performs a reliable local-to-global explanation synthesis. Explanation fidelity is preserved under strong consolidation, the resulting global explanations are structurally stable, and the identified temporal behaviours align with domain-level signal characteristics. Rather than producing arbitrary or fragmented aggregates of importance scores, L2GTX highlights shared, decision-relevant temporal primitives that support meaningful interpretation of time-series classifiers.

\FloatBarrier

\section{Conclusion and Future Work}
\label{sec:conclusion}

In this work, we introduced L2GTX, a model-agnostic method for producing interpretable, class-wise global explanations of time-series classifiers by aggregating local explanations from a selective and representative set of instances. Rather than relying on a single global surrogate, L2GTX builds global understanding directly from faithful instance-level explanations, enabling a principled transition from local reasoning to global model behaviour. This addresses a key limitation in existing time series XAI approaches, which predominantly provide local or architecture-specific explanations and lack principled mechanisms for synthesising faithful class-wise global summaries.

Through experiments on multiple time-series datasets, we show that L2GTX produces compact global explanations while maintaining stable Global Faithfulness. The consistency of faithfulness scores across different levels of explanation consolidation suggests that selecting a diverse and representative set of instances is effective for capturing invariant class-level temporal patterns, while suppressing redundant or instance-specific variation. Moreover, explanations generated across different model architectures exhibit strong structural agreement, indicating that L2GTX highlights shared decision-relevant temporal cues. In contrast to approaches that aggregate raw attribution scores or depend on model-internal representations, L2GTX preserves temporal structure and yields semantically interpretable global patterns.

Future work will extend L2GTX to multivariate time series, where capturing interactions and dependencies across channels introduces additional challenges for global interpretability. Additionally, human-centred evaluation of global explanations, including structured assessments with domain experts, could further examine the perceived usefulness, clarity, and trustworthiness of the generated class-level summaries in real-world decision-making settings.

%
%
%
\bibliographystyle{unsrtnat}
\bibliography{references}
\newpage
\appendix
\section{Additional Experimental Details}
\label{app:details}

\subsection{LOMATCE Algorithm}
\label{app:lomatce}

\begin{algorithm}[H]
\caption{LOMATCE: LOcal Model-Agnostic Time Series Classification Explanations}
\label{alg:lomatce}
\begin{algorithmic}[1]
\State \textbf{Input:} Time series $X$, Black-box $f$, Number of samples $N_z$, Bandwidth $\sigma$
\State \textbf{Output:} Local Cluster Importance Weights $\hat{\beta}$

\State \textbf{Stage A: Neighborhood Generation}
\State $\mathcal{Z} \gets \{X\}$ \Comment{Include original instance}
\For{$i = 1$ \textbf{to} $N_z - 1$}
    \State Choose $method \in \{0, \mu_{\text{seg}}, \mu_{\text{total}}, \text{rnd}\}$
    \State $z \gets \text{Perturb}(X, method)$ \Comment{Apply random segment perturbation}
    \State $\mathcal{Z} \gets \mathcal{Z} \cup \{z\}$
\EndFor

\State \textbf{Stage B: Distance and Weighting}
\For{each $z \in \mathcal{Z}$}
    \State $d \gets \text{FastDTW}(X, z)$
    \State $\pi_{X}(z) \gets \exp(-d^2 / \sigma^2)$
\EndFor

\State \textbf{Stage C: PEP Extraction and Transformation}
\State $\mathcal{E} \gets \text{ExtractPEPs}(\mathcal{Z})$ \Comment{Extracting increasing, decreasing, max and min}
\State $\mathcal{C} \gets \text{KMeans}(\text{flattened } \mathcal{E})$ \Comment{Optimal $k$ via Silhouette}
\State $D \gets \text{EventAttribution}(\mathcal{E}, \mathcal{C})$ \Comment{Build count-based matrix}

\State \textbf{Stage D: Surrogate Training}
\State $Y \gets f(\mathcal{Z})$ \Comment{Get black-box predictions}
\State $\hat{\beta} \gets \text{RidgeRegression}(D, Y, \text{weights}=\pi_{X})$

\State \Return $\hat{\beta}$
\end{algorithmic}
\end{algorithm}







\end{document}